\documentclass[sigconf,screen]{acmart}

\AtBeginDocument{%
  \providecommand\BibTeX{{%
    \normalfont B\kern-0.5em{\scshape i\kern-0.25em b}\kern-0.8em\TeX}}}

\renewcommand\footnotetextcopyrightpermission[1]{} 
\setcopyright{none}

\usepackage{multirow}
\usepackage{colortbl}
\usepackage{bm}
\usepackage{pifont}
\usepackage{hyperref}
\begin{document}

\title{DAPE: Dual-Stage Parameter-Efficient Fine-Tuning for Consistent Video Editing with Diffusion Models}


\author{Junhao Xia}
\email{xiajh23@mails.tsinghua.edu.cn}
\affiliation{%
  \institution{Tsinghua University}
  \city{Beijing}
  \country{China}
}
\authornote{\normalsize Homepage: \href{https://junhaoooxia.github.io/DAPE.github.io/}{https://junhaoooxia.github.io/DAPE.github.io/}}

\author{Chaoyang Zhang}
\email{cy-zhang23@mails.tsinghua.edu.cn}
\affiliation{%
  \institution{Tsinghua University}
  \city{Beijing}
  \country{China}
}

\author{Yecheng Zhang}
\email{zhangyec23@mails.tsinghua.edu.cn}
\affiliation{%
  \institution{Tsinghua University}
  \city{Beijing}
  \country{China}
}

\author{Chengyang Zhou}
\email{chengyang.zhou@duke.edu}
\affiliation{%
  \institution{Duke University}
  \city{Durham}
  \state{NC}
  \country{USA}
}

\author{Zhichang Wang}
\email{wzcc@stu.pku.edu.cn}
\affiliation{%
  \institution{Peking University}
  \city{Shenzhen}
  \country{China}
}

\author{Bochun Liu}
\email{liu-bc23@mails.tsinghua.edu.cn}
\affiliation{%
  \institution{Tsinghua University}
  \city{Beijing}
  \country{China}
}

\author{Dongshuo Yin}
\email{yinds@mail.tsinghua.edu.cn}
\affiliation{%
  \institution{Tsinghua University}
  \city{Beijing}
  \country{China}
}
\authornote{\normalsize Corresponding author.}



\begin{abstract}
Video generation based on diffusion models presents a challenging multimodal task, with video editing emerging as a pivotal direction in this field. Recent video editing approaches primarily fall into two categories: training-required and training-free methods. While training-based methods incur high computational costs, training-free alternatives often yield suboptimal performance. To address these limitations, we propose \textbf{DAPE}, a high-quality yet cost-effective two-stage parameter-efficient fine-tuning (PEFT) framework for video editing. In the first stage, we design an efficient norm-tuning method to enhance temporal consistency in generated videos. The second stage introduces a vision-friendly adapter to improve visual quality. Additionally, we identify critical shortcomings in existing benchmarks, including limited category diversity, imbalanced object distribution, and inconsistent frame counts. To mitigate these issues, we curate a large dataset benchmark comprising 232 videos with rich annotations and 6 editing prompts, enabling objective and comprehensive evaluation of advanced methods. Extensive experiments on existing datasets (BalanceCC, LOVEU-TGVE, RAVE) and our proposed benchmark demonstrate that DAPE significantly improves temporal coherence and text-video alignment while outperforming previous state-of-the-art approaches. 
\end{abstract}




\begin{teaserfigure}
  \centering
  \includegraphics[width=.85\textwidth]{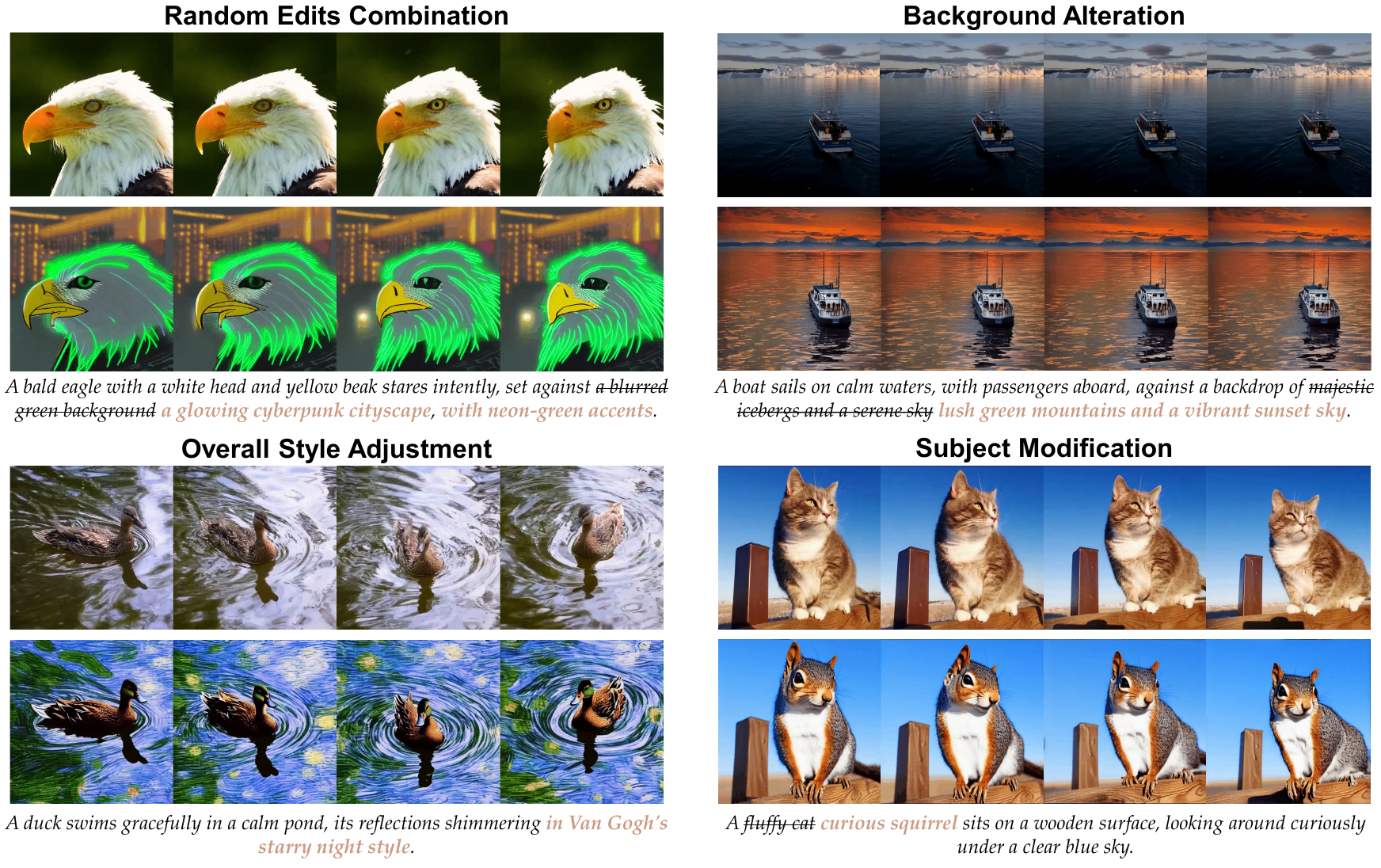}
  \caption{\textbf{DAPE} is a high-quality and cost-effective dual-stage parameter-efficient fine-tuning framework for text-based video editing.
  The diagram presents the performance of our method (lower) on original videos (upper) across four typical scenarios. 
  }
  \label{fig:teaser}
\end{teaserfigure}


\maketitle

\section{Introduction}

Video generation~\cite{vondrick2016generating, VDMho2022video, MAVsinger2022make,imagenvideoho2022,openai2024sora} has emerged as one of the most challenging and promising research directions within computer vision in recent years. As a prominent subfield, video editing~\cite{TAVwu2023, qi2023fatezero, videop2pliu2024,yang2025videograin} aims to controllably modify the visual elements (e.g., objects, backgrounds) semantic information (e.g., textual descriptions), or dynamic characteristics (e.g., motion trajectories) of existing video content while maintaining spatio-temporal coherence.~\cite{sun2024videoeditiingvsurvey} This technique holds significant commercial value, especially in areas such as metaverse and digital human creation, drawing considerable attention from leading technology companies like Microsoft~\cite{feng2024ccedit,xing2024simda}, Google~\cite{imagenvideoho2022}, Nvidia~\cite{blattmann2023align} and OpenAI~\cite{openai2024sora}. Figure \ref{fig:teaser} shows four typical applications of video editing.

Inspired by the recent success of diffusion models~\cite{ho2020ddpm,dhariwal2021diffusion} and image editing methods~\cite{hertz2022p2p,couairon2022diffedit,brooks2023instructpix2pix}, contemporary video editing approaches typically adopt DDIM Inversion~\cite{song2020ddim} to add noise to the target videos and subsequently apply various conditioning strategies during denoising to facilitate content editing. For instance, RAVE~\cite{kara2024rave} enhances temporal consistency via grid concatenation and noise shuffling for conditional injection, while CCEdit~\cite{feng2024ccedit} improves the precise and creative editing capabilities by introducing a novel trident network structure that separates structure and appearance control. However, training-based methods generally incur high computational costs, whereas training-free methods typically struggle to achieve high-quality results. Balancing computational efficiency and video generation quality remains a critical challenge in video editing research~\cite{sun2024videoeditingsurvey}.

In visual tasks~\cite{yin2023lorand,yin2024parameter,yin20245} and text tasks~\cite{hu2022lora, lester2021power}, Parameter-efficient fine-tuning (PEFT) techniques have been widely employed to enhance the performance of large-scale models on specific downstream tasks, such as image recognition~\cite{zhang2020sidetuning} and object segmentation~\cite{peng2024peftinCV}. PEFT methods optimize only a small subset of model parameters, thus significantly reducing training costs and enhancing model performance on downstream tasks even with limited training data~\cite{houlsby2019parameter,xin2024peftvisionmodel}. Video editing tasks based on diffusion models often use a single video template to generate new videos~\cite{TAVwu2023, kara2024rave}, inherently forming a few-shot learning scenario~\cite{song2023fewshotsurvey}. Hence, leveraging PEFT to balance computational cost and video editing quality is highly promising. Despite its potential, PEFT remains under-explored in video editing, and it is essential to conduct a comprehensive investigation into its value within video editing tasks.

To address the challenge of optimizing video editing performance and computational efficiency, we propose DAPE, a novel dual-stage parameter-efficient fine-tuning approach for video editing designed to enhance temporal and visual consistency. First, recent studies have demonstrated that parameters play crucial roles in enhancing conditional control~\cite{peebles2023dit,huang2017adaInsNorm} and visual understanding~\cite{basu2024LN_Tuning}, with recent evidence shows that temporal consistency in text-to-video (T2V) generation is particularly sensitive to normalization scales within temporal layers~\cite{EI2}. To address this sensitivity, we propose a new norm-tuning strategy and introduce a learnable scale factor to balance the original and normalized features optimally. Second, adapter-tuning has been demonstrated to enhance model adaptability for capturing data features effectively, especially in few-shot scenarios~\cite{zhang2022fewshotadapter}. To improve model comprehension of single-video templates, we design a visual adapter module strategically integrated into the diffusion model. In exploring the individual effects of these two optimization schemes, we find that separately, each significantly enhances either temporal consistency or visual quality. However, jointly training them introduces negative interactions, compromising their respective strengths. Consequently, we finally adopt the dual-stage framework to mitigate these adverse effects, as validated by comprehensive ablation studies in Sec.~\ref{Ablation Study}. Furthermore, existing video editing benchmarks suffer from excessive frame lengths, low visual quality, and limited content diversity, thus inadequately assessing overall model capabilities. To address these limitations, we present a novel large-scale dataset, DAPE Dataset, characterized by standardized format, high-quality visuals, and a wide variety of video types. The DAPE Dataset comprises 232 videos, each accompanied by a detailed video caption, video element types annotations, video scene complexity labels, and a set of diverse editing prompts. Extensive experimental evaluations conducted on our DAPE Dataset and three representative benchmarks (RAVE Dataset~\cite{kara2024rave}, BalanceCC~\cite{feng2024ccedit}, loveu-tgve~\cite{wu2023loveu-tgve}) demonstrate that our proposed method quantitatively and qualitatively outperforms previous state-of-the-art, substantially advancing temporal and visual consistency in video editing. 

The key contributions of our work are summarized as follows:
\begin{itemize}
    \item We propose a novel dual-stage parameter-efficient fine-tuning method to significantly improve temporal and spatial consistency in video editing tasks.
    \item We design effective PEFT modules for the video editing tasks during each stage respectively, aiming to optimize temporal consistency and visual feature comprehension.
    \item We introduce a large-scale, high-quality DAPE Dataset, enabling comprehensive and objective assessment of video editing methods.
    \item Extensive experiments on multiple datasets (DAPE Dataset, RAVE Dataset, BalanceCC, loveu-tgve) validate the superior performance of our method, outperforming previous state-of-the-art quantitatively and qualitatively.
\end{itemize}

\section{Related Work}

\subsection{Text-to-Video Generation}

Early approaches to text-to-video generation employed Generative Adversarial Networks (GANs) and primarily focused on domain-specific video synthesis~\cite{xing2024survey}. With the advent of diffusion models, researchers discovered that diffusion-based architecture could also achieve competitive video generation quality. To address the inherent challenges of video data modeling and the scarcity of large-scale, high-quality text-video datasets, pretrained text-to-image (T2I) diffusion models have been adapted by enhancing their spatial-temporal consistency to develop T2V frameworks. For instance, Video Diffusion Models~\cite{VDMho2022video} pioneered the application of diffusion models for video generation, Make-A-Video~\cite{MAVsinger2022make} leveraged the DALL-E 2~\cite{dalle2} architecture to learn cross-frame motion patterns from video data, and Imagen Video~\cite{imagenvideoho2022} extended the Imagen~\cite{imagensaharia2022} framework through joint text-image-video training. Additionally, Video LDM~\cite{videoldm}, Latent Shift~\cite{latentshift}, and VideoFactory~\cite{videofactory}, have utilized open-source Stable Diffusion models as foundational backbones. More recently, advancements in T2V models focused on architectural innovations (e.g., SORA~\cite{openai2024sora}, CogVideoX~\cite{yang2024cogvideox}), video sampling acceleration, and temporal coherence refinement. Despite remarkable progress, training T2V models from scratch remains challenging due to the requirement for large-scale, high-quality text-video pairs (e.g., WebVID-10M~\cite{WebVID10Mbain2022}, MSR-VTT~\cite{xu2016msr}, LAION-5B~\cite{laion-5Bschuhmann2022}) and substantial computational resources.

\subsection{Text-Guided Video Editing}

Text-guided video editing offers an efficient and lightweight alternative for video generation by adapting T2I diffusion models to modify video content while preserving original motion dynamics. This paradigm can be broadly categorized into two approaches, training-based and training-free. Training-based approaches typically fine-tune temporal layers of diffusion models to capture inter-frame temporal relationships. For example, Tune-A-Video~\cite{TAVwu2023} introduced temporal attention for one-shot video synthesis, while Edit-A-Video~\cite{EditAVideoshin2024edit} proposed "sparse-causal blending" to mitigate background inconsistency alongside null text inversion. Video-P2P~\cite{videop2pliu2024} extended prompt-to-prompt editing to videos via shared embedding optimization and cross-attention control. EI2~\cite{EI2} improved temporal coherence through redesigned attention mechanisms. 

Training-free methods, on the contrary, often utilize frame-level feature guidance or auxiliary conditions (e.g., depth maps, sketches) to enhance consistency. The former type includes works such as Tokenflow~\cite{geyer2023tokenflow} which improved temporal alignment by enforcing semantic correspondence in diffusion representations across frames and FateZero~\cite{qi2023fatezero} which preserved attention features during inversion and blended them into the editing process, and the latter contains methods like Render-A-Video~\cite{yang2023rerenderavideo} which employed optical flow to guide hierarchical cross-frame constraints, ControlVideo~\cite{zhang2023controlvideo} which integrated ControlNet with interleaved-frame smoothing as well as RAVE~\cite{kara2024rave}, an approach to enhancing denoising via grid concatenation and noise shuffling for conditional injection. Although training-based methods excel in generalization capacity for novel editing requirements, they incur higher computational costs compared to training-free alternatives.

\subsection{Parameter-Efficient Fine-Tuning}

In natural language processing (NLP), Parameter-Efficient Fine-Tuning (PEFT) techniques alleviate the computational overhead associated with fully fine-tuning models for downstream tasks by reducing the number of trainable parameters while maintaining performance. Recent investigations in video generation have also explored PEFT approaches. For instance, SimDA~\cite{xing2024simda} efficiently adapted a 1.1-billion-parameter text-to-image model for video synthesis using only 24 million trainable parameters. CAMEL~\cite{camelzhang2024} introduces prompt-tuning to summarize motion concepts from videos while ExVideo~\cite{duan2024exvideo} achieved long-video generation by leveraging 3D convolutions and parameter-efficient post-tuning. However, existing research has not yet systematically investigated how PEFT methods influence temporal consistency and text-video alignment, and this work primarily focuses on this topic.

\begin{figure*}
    \centering
    \includegraphics[width=1.0\textwidth]{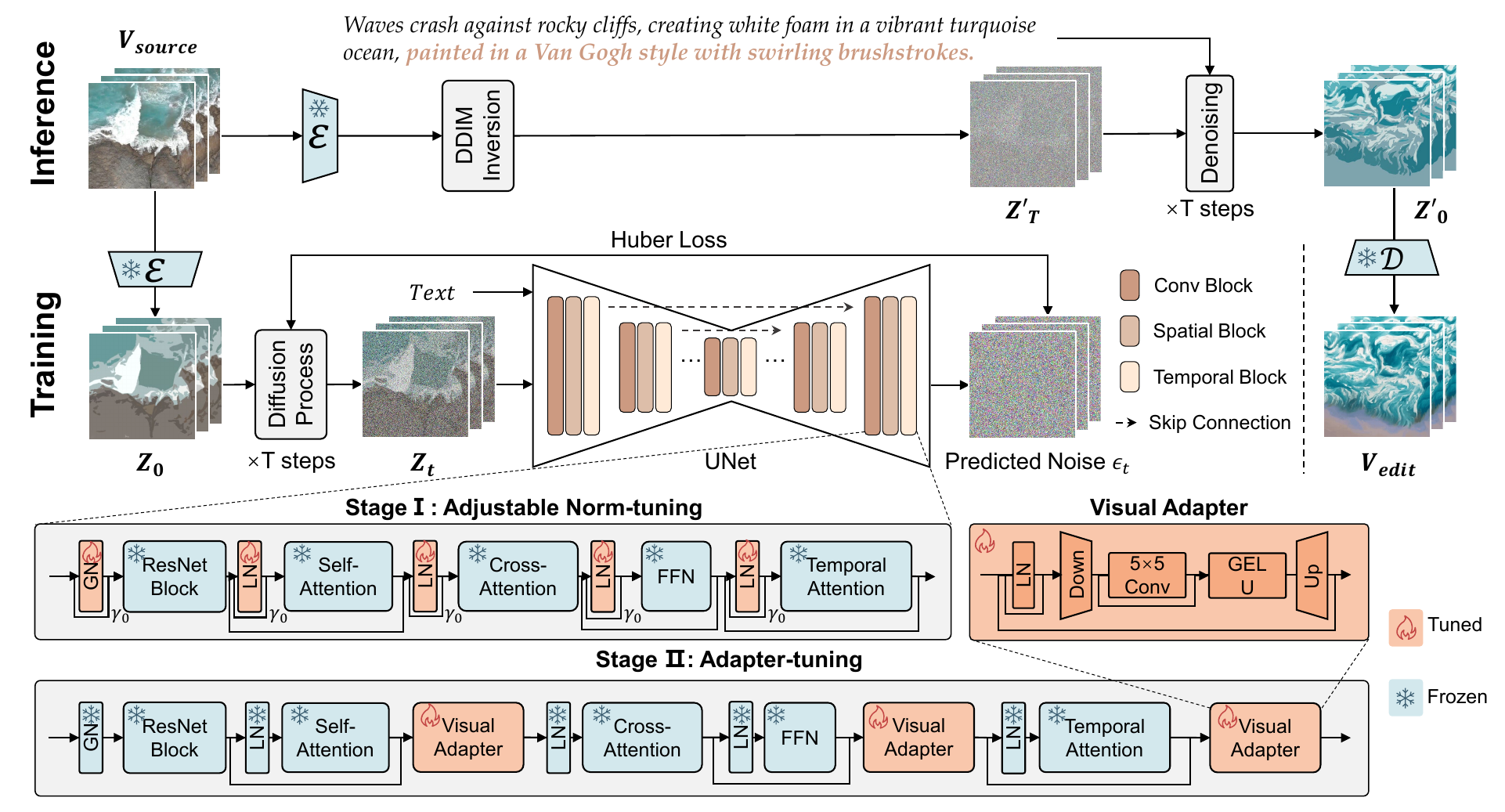}
    \caption{\textit{Overall of DAPE.} DAPE is based on the diffusion model. In the first stage, only the norm layers are fine-tuned. In the second stage, the visual adapter is inserted at specific positions for fine-tuning.}
    \label{fig:pipeline}
\end{figure*}

\section{Methodology}
In this section, we first introduce the fundamental concept in Sec. ~\ref{Preliminaries}, namely latent diffusion models and adapter tuning, which are pivotal to our framework and detailedly demonstrate the DAPE framework in Sec. ~\ref{framework}.

\subsection{Preliminaries}
\label{Preliminaries}

\noindent \textbf{Latent Diffusion Models (LDMs).} LDMs~\cite{rombach2022high} are efficient variants of DDPMs~\cite{ho2020ddpm} that operate the diffusion process in a latent space. They are mainly built upon two key components. First, an auto-encoder maps images $x$ to the latent space $z = \mathcal{E}(x)$ and reconstructs them via $\mathcal{D}(z)$ enabling $\mathcal{D}(\mathcal{E}(x)) \approx x$. The diffusion process is then performed on $z$, using a U-Net based network to predict the added noise $\bm{\epsilon}_\theta$. The objective of LDMs is as follows: 
\begin{equation}
\mathbb{E}_{z, \epsilon \sim \mathcal{N}(0,1), t, c} \left[\|\epsilon - \epsilon_\theta(z_t, t, c)\|^2_2\right],
\end{equation}
where $z_t$ denotes the noisy latent at timestep $t$, and $c$ represents the text condition embedding.

\noindent \textbf{Adapter Tuning.} 
As a typical parameter-efficient fine-tuning method, adapter tuning refers to the approach that integrating small, trainable modules into models and fine-tuning them during training~\cite{houlsby2019parameter}. These learnable structure can facilitate robust performance in specific downstream tasks by capturing domain-specific variations while avoiding catastrophic forgetting. A conventional adapter module can be formulated as follows:
\begin{equation}
\text{Adapter}(\mathbf{X}) = \mathbf{X} + \mathbf{W}_{\mathrm{up}}\left(\phi\left(\mathbf{W}_{\mathrm{down}}(\mathbf{X})\right)\right),
\end{equation}
where $\mathbf{W}_{\text{down}}$ and $\mathbf{W}_{\text{up}}$ are the learnable projection matrices, and $\phi(\cdot)$ denotes an activation function.

\subsection{Framework}
\label{framework}

In this section, we demonstrate the framework of our proposed DAPE. It is a diffusion-based dual-stage parameter-efficient fine-tuning approach to generate consistent videos with high quality.

\noindent \textbf{DAPE Architecture.}
As illustrated in Figure~\ref{fig:pipeline}, DAPE learns cross-frame temporal features via adjustable norm-tuning and captures local visual features by visual adapter from a single video. It adopts a dual-stage paradigm that decouples the learning of temporal and visual features to effectively mitigating strength conflicts, as is supported by ablation studies. During inference, same as most video editing approaches~\cite{TAVwu2023,kara2024rave,geyer2023tokenflow}, it uses DDIM Inversion~\cite{song2020ddim} to retain the original video's features within the initial noise and progressively removes the U-Net-predicted noise conditioned on various inputs, ultimately generating the edited video.

\noindent \textbf{Adjustable norm-tuning.}
Motivated by recent findings highlighting the pivotal role of normalization layers in shaping the quality and consistency of generation~\cite{peebles2023dit,EI2}, we introduce a novel approach, namely adjustable norm-tuning, to optimize normalization parameters of diffusion models blocks including ResNet blocks and attention blocks. To further enhance the model’s adaptability, a learnable affine parameters $\gamma_0$ is incorporated in the norm-tuning step. $\gamma_0$ is initialized to 0 as conventional normalization conduct and is multiplied on latent representations $z_t$. In the lower part of Figure~\ref{fig:pipeline}, stage I can be formulated as follows:
 
\begin{equation}
    \hat{z}_t = \gamma \cdot Norm(z_t) + \beta + \gamma_0 \cdot z_t,
\end{equation}
where $z_t$ is the input latent feature at timestep $t$, $Norm(\cdot)$ denotes a normalization operation with learnable parameters $\gamma, \beta$.

\noindent \textbf{Visual Adapter.}
Adapters have been widely used to capture visual features in image tasks~\cite{yin2023lorand}. To improve the stability of training and model adaptability, a layer normalization block with a learnable scaling parameters $w_0$ is adopted, followed by down projection, convolution layer, nonlinear activation, up projection, and skip connections. Notably, to enhance spatial perceptual capabilities while minimizing additional parameters, convolution layer using a single depth-wise $5 \times 5$ kernel, leading to measurable improvements in extensive experiments. The procedure can be formally described as follows, also shown in stage II from Figure~\ref{fig:pipeline}:
\begin{align}
    z &= z_0 + Up\left(\sigma\left(f\left(Down(z_{\text{norm}})\right)\right)\right), \\
    f &= z + \omega_{dw} \otimes_{dw} z_{down},
\end{align}
where $\sigma$ is the activation function, $z_{down}$ represents the down-sampled version of $z_{norm}$, $\omega_{dw}$ denotes the convolutional kernel and $\otimes_{dw}$ indicates depth-wise convolution.

\noindent \textbf{Position Consideration.}
The effects of incorporating visual adapters into different layers of the model intrigued our interest. Our ablations reveal that integrating the visual adapter exclusively within the first cross-attention block of the up-sampling (decoder) layers yields the best performance both in temporal coherence and alignment while saving the parameter size. Therefore, we adopt this insertion position in the DAPE architecture.

\noindent \textbf{Loss Function.}
While mean squared error (MSE) loss is a common choice for diffusion-based generative models, it is vulnerable to outliers in training data. Considering the distribution discrepancy between the pretrained dataset domain and individual video samples, we adopt huber loss as loss function, which combines the robustness of L1 loss with the stability of MSE. The Huber loss is defined as:
\begin{equation}
\mathcal{L}_{\delta}(r) = 
\begin{cases}
\frac{1}{2} r^2 & \text{if } |r| \leq \delta, \\
\delta \cdot (|r| - \frac{1}{2}\delta) & \text{otherwise},
\end{cases}
\end{equation}
where $r$ is the residual between the predicted and target noise, and $\delta$ is a threshold hyperparameter.

\section{DAPE Benchmark}

\subsection{Establishment}
Despite the availability of several datasets in the field of video editing, current benchmarks still suffer from key limitations, including inconsistent resolution and frame count, low visual quality such as excessive camera motion and image blur, and limited content diversity. These flaws hinder researchers from conducting comprehensive and objective assessments of model performance. To mitigate evaluation bias caused by dataset limitations, we introduce the \textbf{DAPE Dataset}, a standardized benchmark specifically designed to support high-quality and content-diverse video editing tasks.

We initially collected about 2,134 videos from the multiple sources ~\cite{kara2024rave, xu2016msr, wang2019vatex, pixabay},  all of which are authorized for commercial use. To construct a high-quality video dataset, we apply a three-step processing pipeline: 1) \textbf{Standardization}, where all videos are resized to a fixed resolution of 512×512 and trimmed to a standardized number of frames (32, 64, or 128); 2) \textbf{Automatic filtering}, which involves optical flow analysis to remove videos with excessive motion and scene cut detection to exclude those containing temporal discontinuities in line with previous work ~\cite{blattmann2023stable, wang2024egovid}; and 3) \textbf{Manual verification}, where the remaining videos are reviewed by annotators to ensure clarity, stability, and overall visual quality. Adapted from MSR-VTT ~\cite{xu2016msr}, we decompose each video into three core components: subjects, backgrounds, and events. During the selection process, we deliberately ensured diversity across these components in both categorical coverage and complexity levels, leading to a relatively balanced distribution. As a result, we obtained a curated set of 232 high-quality videos with broad content coverage. 

Following prior work~ \cite{teodoro2024mive, feng2024ccedit}, we employed the state-of-the-art Qwen2.5 vision-language model ~\cite{Qwen2.5-VL} to generate textual captions and assign a complexity level to each video. Based on these captions, GPT-4o ~\cite{hurst2024gpt} was used to generate diverse prompts tailored to video editing tasks. All generated captions, complexity annotations, and prompts were also manually reviewed to ensure their accuracy.

\subsection{Statistics}

\begin{figure}
    \centering
    \includegraphics[width=\columnwidth]{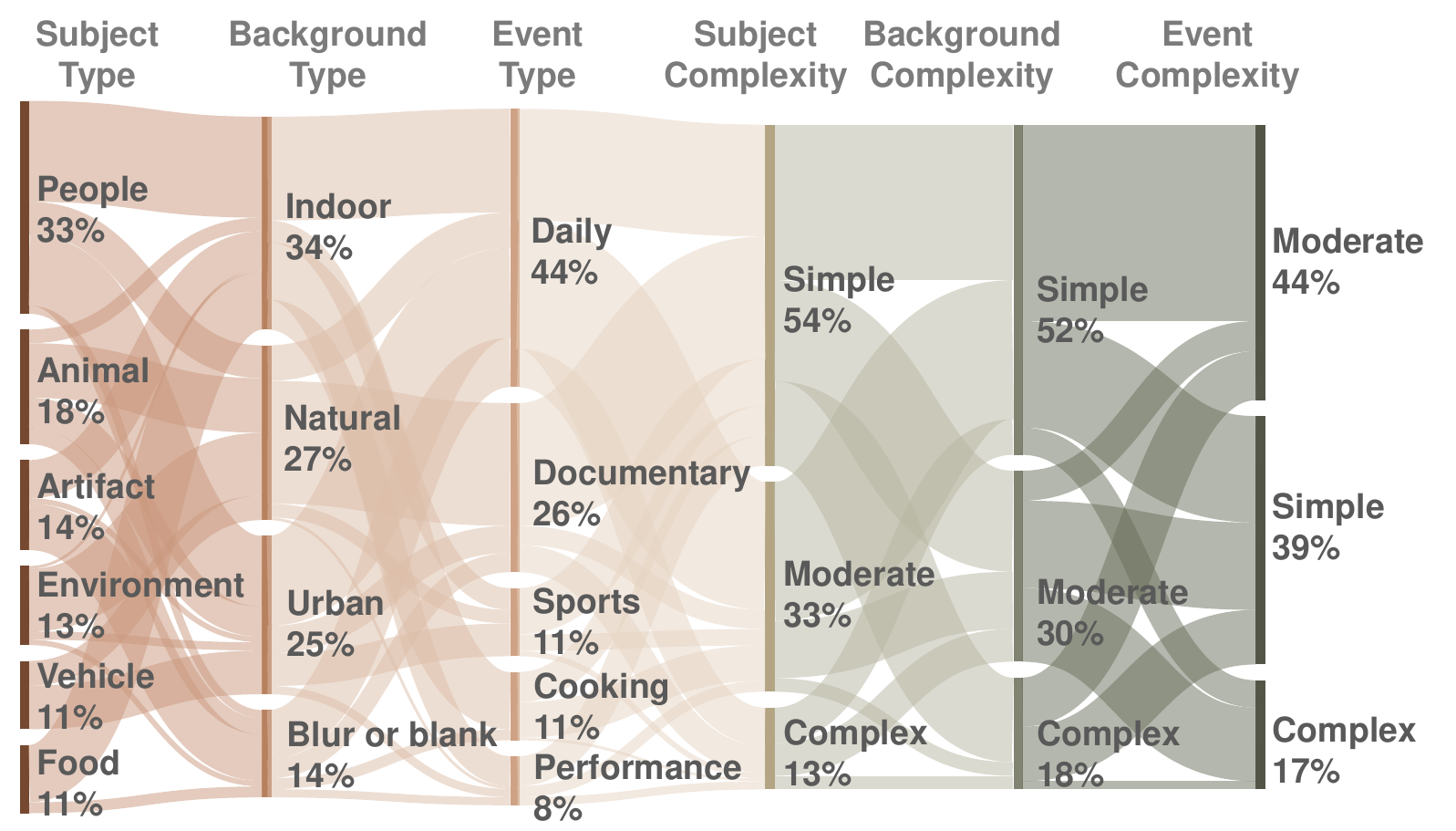}
    \caption{\textit{Dataset statistics. }Distributions of the DAPE Dataset across six semantic dimensions: category and complexity for subject, background, and event.}
    \label{fig:statistics}
    \vspace{-10pt}
\end{figure}

The overall distribution of semantic categories and complexity levels in the DAPE Dataset is illustrated in Figure~\ref{fig:statistics}.

For subject type, the “people” category is the most prevalent (33\%), followed by “animal” (18\%), while “artifact,” “environment,” “vehicle,” and “food” collectively make up the remainder. This designed choice reflects the dominance of human-centric content in real-world video scenarios. Regarding background and event types, the distribution is relatively balanced. Indoor scenes appear most frequently (34\%), and “daily” events are the most common (44\%), aligning with the characteristics of everyday user-generated content. Each of the three components is further annotated with a three-level complexity score: \textit{simple, moderate, and complex}. The dataset is intentionally constructed to emphasize simple and moderate levels across all dimensions, considering the current maturity of video editing models. 

Due to the page limit, we provide details of the dataset annotation process in the supplementary material, including: video source, video annotation, prompt generation, video selection criteria, video categories, types of edits, visualizations of dataset samples.
\section{Experiments}
\label{experiment}

\subsection{Settings}
\noindent \textbf{Implementation Details.}
DAPE employs the pre-trained T2I model, stable diffusion-v1.5, along with the temporal layers from CCEdit~\cite{feng2024ccedit} as initialization weights. Adjustable norm-tuning stage employs 400 timesteps with a learning rate of $5 \times 10^{-5}$ and a batch size of 1, while the visual adapter tuning stage involves 70 timesteps at a learning rate of $1 \times 10^{-5}$ with the same batch size. During inference, we set the DDIM~\cite{song2020ddim} sampler configured for 50 steps, classifier-free guidance~\cite{cfg} with a strength factor of 7.5. Besides, we use pre-trained ControlNet structure~\cite{feng2024ccedit} for additional condition during inference. Our experiments are conducted on 8 NVIDIA A800 GPUs.

\noindent \textbf{Baselines.}
We select five latest baseline methods covering both training-based and training-free approaches using their official implementations, including \textbf{Tune-A-Video (ICCV'23)} \cite{TAVwu2023}, \textbf{CAMEL (CVPR'24)} \cite{camelzhang2024}, \textbf{SimDA (CVPR'24)} \cite{xing2024simda}, \textbf{RAVE (CVPR'24)} \cite{kara2024rave}, and \textbf{CCEdit (CVPR'24)} \cite{feng2024ccedit}. Our proposed DAPE framework can also be applied to other frameworks. Therefore, we conducted many experiments based on each baseline to demonstrate the potential insights and implications of our approach for other models.

\noindent \textbf{Datasets.}
To fully demonstrate the effectiveness of our methods, we conduct experiments on the proposed \textbf{DAPE dataset} and three other lastest and typical video editing datasets: \textbf{1) LOVEU-TGVE~\cite{wu2023loveu-tgve}:} 76 videos selected from DAVIS~\cite{perazzi2016DAVIS}, Youtube~\cite{youtube} and Videvo~\cite{videvo} with 304 text-video pairs. Each video consists of either 32 or 128 frames, with a resolution of $480 \times 480$. \textbf{2) RAVE Dataset~\cite{kara2024rave}:} 31 videos from diverse sources including Pexels~\cite{pexels}, Pixabay~\cite{pixabay}, and DAVIS~\cite{perazzi2016DAVIS}, with 186 text-video pairs. The video lengths are classified into 8, 36, and 90 frames, with resolutions of $512 \times 512$, $512 \times 320$, or $512 \times 256$. \textbf{3) BalanceCC~\cite{feng2024ccedit}:} 100 open-license videos with a uniform resolution of $512 \times 512$. Each video has 4 edited prompts and the number of video frames ranges from 8 to 1627.

\noindent \textbf{Evaluation Metrics.}
Following established practices in video editing research~\cite{li2024vidtome,TAVwu2023}, we evaluate generated videos primarily from two perspectives: temporal consistency and text-video alignment. 1) \textbf{Temporal consistency}: it consists of CLIP-Frame, which calculates the average pairwise similarity among CLIP~\cite{radford2021CLIP} image embeddings across frames, Interpolation Error and PSNR, computed by interpolating a target frame using adjacent two frames and measuring the error and PSNR between interpolated and source frames to reflect intrinsic video continuity~\cite{jiang2018interpolateerror} and Warping Error~\cite{lai2018wraperror}, which employs RAFT~\cite{teed2020raft} to estimate optical flow between consecutive frames in the original video, and warps edited frames to the next for error computation. 2) \textbf{Text-video alignment}: we use the widely adopted metric CLIP-Text to assess text-video alignment, computing the mean similarity between video frame embeddings and textual embeddings via the CLIP model~\cite{radford2021CLIP}. 

\begin{table*}
\centering

\renewcommand{\arraystretch}{1} 

\scalebox{1}{\setlength{\tabcolsep}{1mm}{
\begin{tabular}{l|ccccc|ccccc}

\hline
\multirow{4}{*}{\centering \textbf{Method}}  & \multicolumn{5}{c|}{\textbf{BalanceCC}} & \multicolumn{5}{c}{\textbf{loveu-tgve}} \\
\cline{2-11}

 & \multicolumn{4}{c|}{\textbf{Temporal Consistency}} & \multicolumn{1}{c|}{\textbf{Alignment}}  & \multicolumn{4}{c|}{\textbf{Temporal Consistency}} & \multicolumn{1}{c}{\textbf{Alignment}} \\
\cline{2-11}

 & CLIP-F $\uparrow$ & Int. Err. $\downarrow$ & Int. PSNR $\uparrow$ & \multicolumn{1}{c|}{War. Err. $\downarrow$} & \multicolumn{1}{c|}{CLIP-T $\uparrow$} & CLIP-F $\uparrow$ & Int. Err. $\downarrow$ & Int. PSNR $\uparrow$ & \multicolumn{1}{c|}{War. Err. $\downarrow$}  & \multicolumn{1}{c}{CLIP-T $\uparrow$} \\

 & $\times$10$^{-2}$ & $\times$10$^{-2}$ & $\times$1 & \multicolumn{1}{c|}{$\times$10$^{-2}$} & \multicolumn{1}{c|}{$\times$10$^{-2}$} & $\times$10$^{-2}$ & $\times$10$^{-2}$ & $\times$1 & \multicolumn{1}{c|}{$\times$10$^{-2}$} & \multicolumn{1}{c}{$\times$10$^{-2}$}\\
\hline
\multicolumn{11}{c}{\textbf{Baselines}} \\
\hline

TAV\cite{TAVwu2023}       & 93.11 & 14.43 & 17.57 & \multicolumn{1}{c|}{5.57}  & \multicolumn{1}{c|}{31.82}      & 4.44 & 10.36 & 20.82 & \multicolumn{1}{c|}{4.05}  & \multicolumn{1}{c}{29.95}   \\
CAMEL\cite{camelzhang2024}              & 94.67 & 8.80  & 22.61 & \multicolumn{1}{c|}{4.07}  & \multicolumn{1}{c|}{29.27}      & 94.44 & 10.14 & 21.11 & \multicolumn{1}{c|}{4.03}  & \multicolumn{1}{c}{27.70}   \\
SimDA\cite{xing2024simda}              & 91.32 & 12.79 & 18.57 & \multicolumn{1}{c|}{5.06}  & \multicolumn{1}{c|}{31.28}      & 91.96 & 9.08  & 21.75 & \multicolumn{1}{c|}{3.11}  & \multicolumn{1}{c}{29.33}   \\
RAVE\cite{kara2024rave}               & 94.10 & 8.69  & 22.05 & \multicolumn{1}{c|}{\underline{2.46}}  & \multicolumn{1}{c|}{\underline{32.15}}      & 94.32 & 8.27  & 22.59 & \multicolumn{1}{c|}{\underline{2.34}}  & \multicolumn{1}{c}{\underline{30.18}}   \\
CCEdit\cite{feng2024ccedit}             & \underline{95.50} & \underline{7.29}  & \underline{24.33} & \multicolumn{1}{c|}{4.52}  & \multicolumn{1}{c|}{29.76}      & 94.00 & \underline{7.65}  & \underline{23.76} & \multicolumn{1}{c|}{3.24}  & \multicolumn{1}{c}{28.80}   \\

\hline
\multicolumn{11}{c}{\textbf{Ours}} \\
\hline

DAPE (TAV)       & 93.46 & 13.98 & 17.83 & \multicolumn{1}{c|}{5.31}  & \multicolumn{1}{c|}{31.82}      & \underline{94.53} & 10.28 & 20.88 & \multicolumn{1}{c|}{3.96}  & \multicolumn{1}{c}{29.98}   \\
\rowcolor{gray!15}
$\Delta_{TAV}$  & \textit{+0.38\%} & \textit{+3.12\%} & \textit{+1.48\%} & \multicolumn{1}{c|}{\textit{+4.67\%}} & \multicolumn{1}{c|}{\textit{0.00\%}}      & \textit{+0.10\%} & \textit{+0.77\%} & \textit{+0.29\%} & \multicolumn{1}{c|}{\textit{+2.22\%}}  & \multicolumn{1}{c}{\textit{+0.10\%}}   \\
DAPE (CAMEL)     & 94.75 & 8.68  & 22.75 & \multicolumn{1}{c|}{3.94}  & \multicolumn{1}{c|}{29.37}      & \textbf{94.67} & 10.04 & 21.20 & \multicolumn{1}{c|}{4.07}  & \multicolumn{1}{c}{27.78}   \\
\rowcolor{gray!15}
$\Delta_{CAMEL}$  & \textit{+0.08\%} & \textit{+1.36\%} & \textit{+0.62\%} & \multicolumn{1}{c|}{\textit{+3.19\%}} & \multicolumn{1}{c|}{\textit{+0.34\%}}      & \textit{+0.24\%} & \textit{+0.99\%} & \textit{+0.43\%} & \multicolumn{1}{c|}{\textit{-0.99\%}} & \multicolumn{1}{c}{\textit{+0.29\%}}   \\
DAPE (SimDA)     & 91.43 & 12.29 & 18.85 & \multicolumn{1}{c|}{4.92}  & \multicolumn{1}{c|}{31.37}      & 92.07 & 8.91  & 21.96 & \multicolumn{1}{c|}{3.01}  & \multicolumn{1}{c}{29.17}   \\
\rowcolor{gray!15}
$\Delta_{SimDA}$  & \textit{+0.12\%} & \textit{+3.91\%} & \textit{+1.51\%} & \multicolumn{1}{c|}{\textit{+2.77\%}} & \multicolumn{1}{c|}{\textit{+0.29\%}}      & \textit{+0.12\%} & \textit{+1.87\%} & \textit{+0.97\%} & \multicolumn{1}{c|}{\textit{+3.22\%}} & \multicolumn{1}{c}{\textit{-0.55\%}}   \\
DAPE (RAVE)      & 94.61 & \textbf{7.18}  & 23.91 & \multicolumn{1}{c|}{\textbf{2.13}}  & \multicolumn{1}{c|}{\textbf{32.85}}      & 94.33 & 7.73  & 23.16 & \multicolumn{1}{c|}{\textbf{2.18}}  & \multicolumn{1}{c}{\textbf{30.35}}   \\
\rowcolor{gray!15}
$\Delta_{RAVE}$  & \textit{+0.54\%} & \textit{+17.38\%} & \textit{+8.44\%} & \multicolumn{1}{c|}{\textit{+13.41\%}} & \multicolumn{1}{c|}{\textit{+2.18\%}}      & \textit{+0.01\%} & \textit{+6.53\%} & \textit{+2.52\%} & \multicolumn{1}{c|}{\textit{+6.84\%}} & \multicolumn{1}{c}{\textit{+0.56\%}}   \\
DAPE (CCEdit)    & \textbf{95.54} & 7.58  & \textbf{24.38} & \multicolumn{1}{c|}{4.03}  & \multicolumn{1}{c|}{30.19}      & 93.76 & \textbf{7.59}  & \textbf{23.85} & \multicolumn{1}{c|}{2.97}  & \multicolumn{1}{c}{29.32}   \\
\rowcolor{gray!15}
$\Delta_{CCEdit}$  & \textit{+0.04\%} & \textit{-3.98\%} & \textit{+0.21\%} & \multicolumn{1}{c|}{\textit{+10.84\%}} & \multicolumn{1}{c|}{\textit{+1.44\%}}      & \textit{-0.26\%} & \textit{+0.78\%} & \textit{+0.38\%} & \multicolumn{1}{c|}{\textit{+8.33\%}} & \multicolumn{1}{c}{\textit{+1.81\%}}   \\

\hline
\end{tabular}%
}}
\vspace{.3em}

\scalebox{1}{\setlength{\tabcolsep}{1mm}{
\begin{tabular}{l|ccccc|ccccc}
\hline
 \multirow{4}{*}{\centering \textbf{Method}} & \multicolumn{5}{c|}{\textbf{RAVE Dataset}} & \multicolumn{5}{c}{\textbf{DAPE Dataset}} \\
\cline{2-11}
 & \multicolumn{4}{c|}{\textbf{Temporal Consistency}} & \multicolumn{1}{c|}{\textbf{Alignment}} &  \multicolumn{4}{c|}{\textbf{Temporal Consistency}} & \multicolumn{1}{c|}{\textbf{Alignment}} \\
\cline{2-11}
 & CLIP-F $\uparrow$ & Int. Err. $\downarrow$ & Int. PSNR $\uparrow$ & \multicolumn{1}{c|}{War. Err. $\downarrow$} & \multicolumn{1}{c|}{CLIP-T $\uparrow$} & CLIP-F $\uparrow$ & Int. Err. $\downarrow$ & Int. PSNR $\uparrow$ & \multicolumn{1}{c|}{War. Err. $\downarrow$}  & \multicolumn{1}{c}{CLIP-T $\uparrow$} \\
 & \(\times 10^{-2}\) & \(\times 10^{-2}\) & \(\times 1\) & \multicolumn{1}{c|}{\(\times 10^{-2}\)} & \multicolumn{1}{c|}{\(\times 10^{-2}\)}  & \(\times 10^{-2}\) & \(\times 10^{-2}\) & \(\times 1\) & \multicolumn{1}{c|}{\(\times 10^{-2}\)} & \multicolumn{1}{c}{\(\times 10^{-2}\)} \\
\hline
\multicolumn{11}{c}{\textbf{Baselines}} \\
\hline
TAV\cite{TAVwu2023}      & 94.35 & 15.03 & 16.64 & \multicolumn{1}{c|}{5.55}  & \multicolumn{1}{c|}{\underline{31.09}}      & 94.88 & 9.00  & 21.73 & \multicolumn{1}{c|}{2.73}  & \multicolumn{1}{c}{31.34}   \\
CAMEL\cite{camelzhang2024}             & 92.85 & 14.18 & 17.36 & \multicolumn{1}{c|}{5.66}  & \multicolumn{1}{c|}{27.40}      & 95.74 & 6.78  & 24.53 & \multicolumn{1}{c|}{2.28}  & \multicolumn{1}{c}{29.95}   \\
SimDA\cite{xing2024simda}             & 91.94 & 13.75 & 17.43 & \multicolumn{1}{c|}{5.40}  & \multicolumn{1}{c|}{30.07}      & 92.22 & 7.96  & 22.75 & \multicolumn{1}{c|}{2.42}  & \multicolumn{1}{c}{30.61}   \\
RAVE\cite{kara2024rave}              & \underline{94.85} & 8.71  & 21.94 & \multicolumn{1}{c|}{\underline{2.53}}  & \multicolumn{1}{c|}{29.76}      & 95.80 & 6.65  & 24.09 & \multicolumn{1}{c|}{\underline{1.37}}  & \multicolumn{1}{c}{\underline{32.52}}   \\
CCEdit\cite{feng2024ccedit}            & 93.74 & 10.34 & 20.37 & \multicolumn{1}{c|}{4.46}  & \multicolumn{1}{c|}{26.41}      & \underline{96.47} & \underline{5.41}  & \underline{26.66} & \multicolumn{1}{c|}{1.90}  & \multicolumn{1}{c}{28.56}   \\
\hline
\multicolumn{11}{c}{\textbf{Ours}} \\
\hline

DAPE (TAV)       & 94.53 & 14.77 & 16.78 & \multicolumn{1}{c|}{5.40}  & \multicolumn{1}{c|}{\textbf{31.19}}      & 94.92 & 9.14  & 21.80 & \multicolumn{1}{c|}{2.66}  & \multicolumn{1}{c}{31.47}   \\
\rowcolor{gray!15}
$\Delta_{TAV}$  & \textit{+0.19\%} & \textit{+1.73\%} & \textit{+0.84\%} & \multicolumn{1}{c|}{\textit{+2.70\%}} & \multicolumn{1}{c|}{\textit{+0.32\%}}      & \textit{+0.04\%} & \textit{-1.56\%} & \textit{+0.32\%} & \multicolumn{1}{c|}{\textit{+2.56\%}} & \multicolumn{1}{c}{\textit{+0.41\%}}   \\
DAPE (CAMEL)     & 92.93 & 14.10 & 17.43 & \multicolumn{1}{c|}{5.47}  & \multicolumn{1}{c|}{27.41}      & 95.89 & 6.61  & 24.74 & \multicolumn{1}{c|}{2.21}  & \multicolumn{1}{c}{30.00}   \\
\rowcolor{gray!15}
$\Delta_{CAMEL}$  & \textit{+0.09\%} & \textit{+0.56\%} & \textit{+0.40\%} & \multicolumn{1}{c|}{\textit{+3.36\%}} & \multicolumn{1}{c|}{\textit{+0.04\%}}      & \textit{+0.16\%} & \textit{+2.51\%} & \textit{+0.86\%} & \multicolumn{1}{c|}{\textit{+3.07\%}} & \multicolumn{1}{c}{\textit{+0.17\%}}   \\
DAPE (SimDA)     & 92.05 & 13.62 & 17.57 & \multicolumn{1}{c|}{5.26}  & \multicolumn{1}{c|}{30.13}      & 93.11 & 7.74  & 23.23 & \multicolumn{1}{c|}{2.37}  & \multicolumn{1}{c}{31.15}   \\
\rowcolor{gray!15}
$\Delta_{SimDA}$  & \textit{+0.12\%} & \textit{+0.95\%} & \textit{+0.80\%} & \multicolumn{1}{c|}{\textit{+2.59\%}} & \multicolumn{1}{c|}{\textit{+0.20\%}}      & \textit{+0.97\%} & \textit{+2.76\%} & \textit{+2.11\%} & \multicolumn{1}{c|}{\textit{+2.07\%}} & \multicolumn{1}{c}{\textit{+1.76\%}}   \\
DAPE (RAVE)      & \textbf{94.98} & \textbf{8.34}  & \underline{22.30} & \multicolumn{1}{c|}{\textbf{2.42}}  & \multicolumn{1}{c|}{29.78}      & 95.85 & 6.27  & 24.63 & \multicolumn{1}{c|}{\textbf{1.26}}  & \multicolumn{1}{c}{\textbf{32.61}}   \\
\rowcolor{gray!15}
$\Delta_{RAVE}$  & \textit{+0.14\%} & \textit{+4.25\%} & \textit{+1.64\%} & \multicolumn{1}{c|}{\textit{+4.35\%}} & \multicolumn{1}{c|}{\textit{+0.07\%}}      & \textit{+0.05\%} & \textit{+5.71\%} & \textit{+2.24\%} & \multicolumn{1}{c|}{\textit{+8.03\%}} & \multicolumn{1}{c}{\textit{+0.28\%}}   \\
DAPE (CCEdit)    & 93.83 & \underline{8.47}  & \textbf{22.37} & \multicolumn{1}{c|}{2.90}  & \multicolumn{1}{c|}{28.35}      & \textbf{96.59} & \textbf{5.31}  & \textbf{27.09} & \multicolumn{1}{c|}{1.52}  & \multicolumn{1}{c}{29.07}   \\
\rowcolor{gray!15}
$\Delta_{CCEdit}$  & \textit{+0.10\%} & \textit{+18.09\%} & \textit{+9.82\%} & \multicolumn{1}{c|}{\textit{+34.98\%}} & \multicolumn{1}{c|}{\textit{+7.35\%}}      & \textit{+0.12\%} & \textit{+1.85\%} & \textit{+1.61\%} & \multicolumn{1}{c|}{\textit{+20.00\%}} & \multicolumn{1}{c}{\textit{+1.79\%}}   \\

\hline
\end{tabular}%
}}
\caption{\textit{\textbf{Quantitative comparison.}} Experiments are conducted on four datasets to evaluate the models' performance on five metrics (CLIP-Frame (CLIP-F), Interpolation Error (Int. Err.), Interpolation PSNR (Int. PSNR), WarpError (War. Err.), CLIP-Text (CLIP-T)). $\uparrow$ means higher is better while $\downarrow$ donates the lower the better. The best and the second-best performance are highlighted in \textbf{bold} and in \underline{underline}, respectively.}
\label{tab:mainresults}
\vspace{-10pt}
\end{table*}

\begin{figure*}
    \centering
    \includegraphics[width=1.0\textwidth]{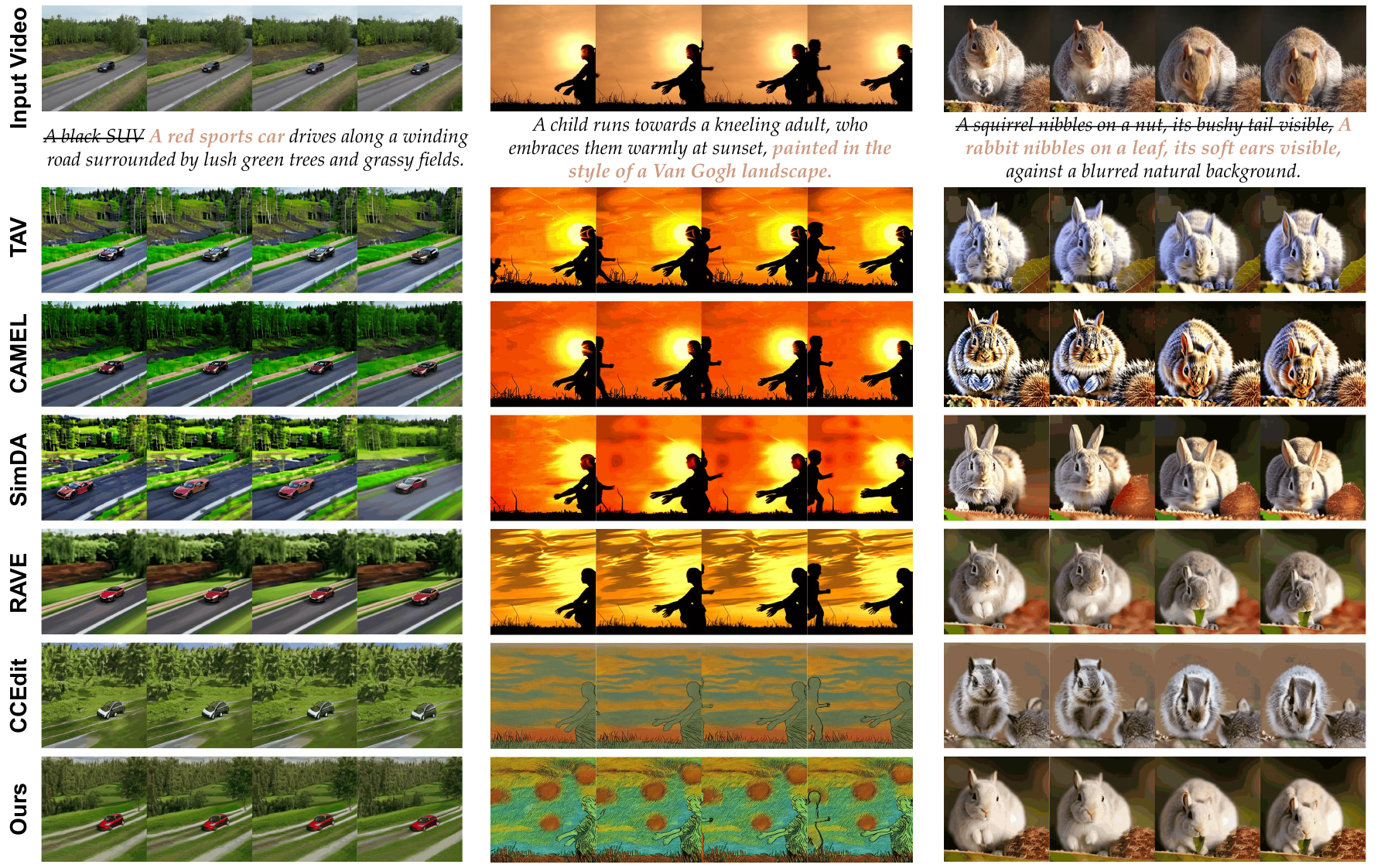}
    \caption{\textit{Qualitative comparison.} Different model performance on given video editing tasks. Our method achieves the best performance in terms of temporal consistency, text alignment and visual quality.}
    \label{fig:qualitativeResults}
\end{figure*}

\vspace{-10pt}
\subsection{Main Results}

\noindent \textbf{Quantitative Results.}
Table~\ref{tab:mainresults} presents the quantitative results of all methods on the four datasets. Macroscopically, DAPE achieves the best performance (highlighted in bold) across all datasets, demonstrating its effectiveness in enhancing the quantitative performance of mainstream video editing tasks. Microscopically, DAPE significantly improves the performance of baseline methods on most metrics, with the highest improvement reaching 34.98\%. These results confirm that the proposed adjustable norm tuning and visual adapter components , as integral elements of our framework, effectively enhance temporal consistency and alignment. Notably, RAVE and CCEdit demonstrate particularly significant reductions in Interpolation Error and Warp Error, and we attribute these improvements to the unique architectural designs of the respective models, since RAVE applies grid-based denoising process while CCEdit refines the appearance and structural guidance in implementation. Additionally, our analysis of the results on the three existing datasets reveals that the ranking of baseline methods varies across different datasets. This observation further underscores the necessity of establishing a large-scale benchmark dataset.


\begin{figure}[h]
    \centering
    \includegraphics[width=1\columnwidth]{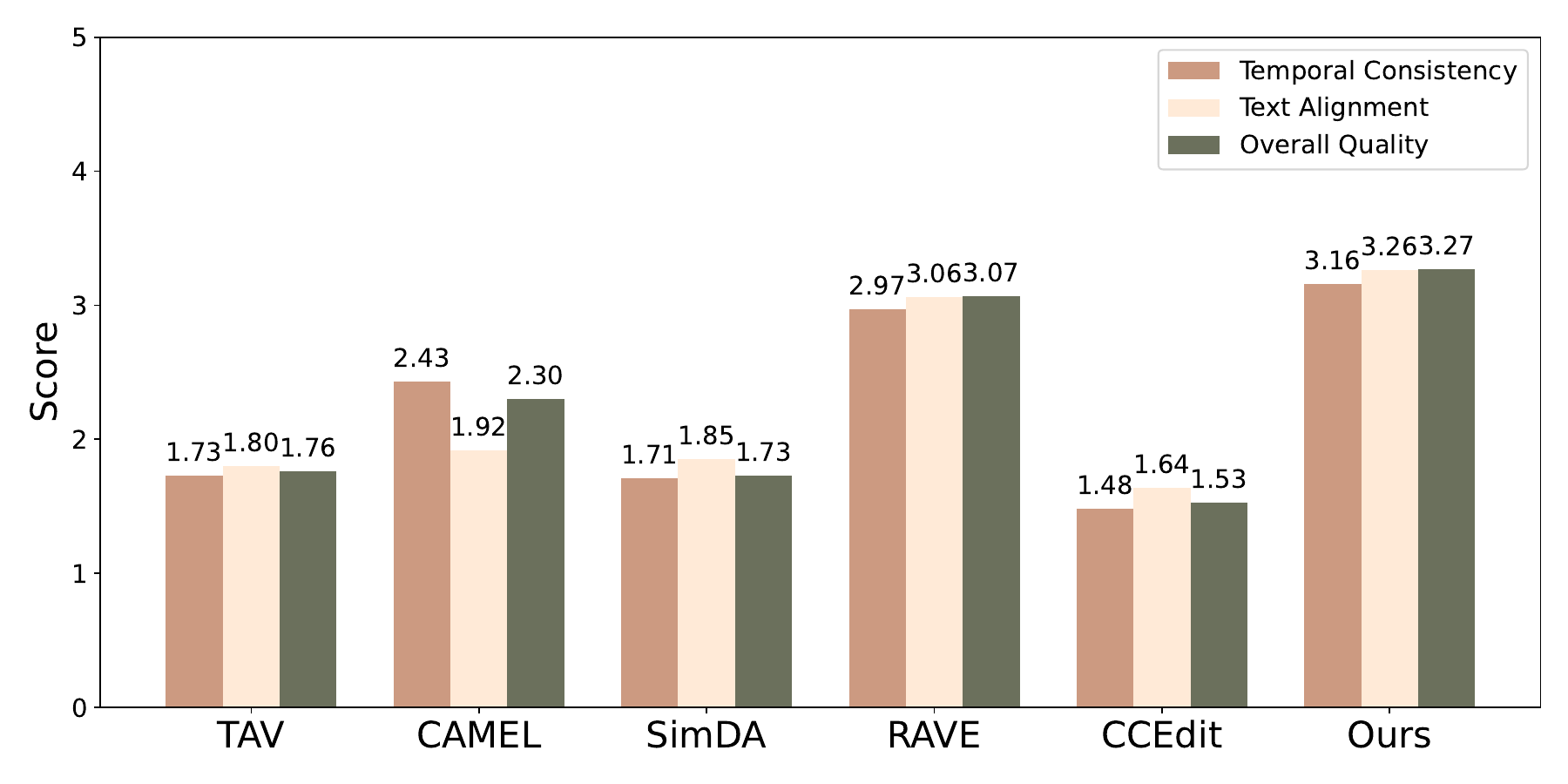}
    \caption{\textit{User Study Results}. Comparison of subjective scores for each model on Temporal Consistency, Text Alignment, and Overall Quality. Models performing the best, second best and third best scores 6, 5 and 4, and the scores for each model are weighted by vote frequency. Our model achieved the highest rating in all three aspects.}
    \label{fig:userstudy}
\end{figure}

\noindent \textbf{Qualitative Results.}
Figure~\ref{fig:qualitativeResults} illustrates the marked differences among the methods regarding temporal consistency, text alignment, and detail quality. The first column needs to change the black SUV to a red sports car. TAV generates a black sports car, CAMEL generates a dark red car, SimDA and RAVE generate low-quality videos, CCEdit generates a non-red car, while DAPE produces a high-quality video with a red sports car. The second column requires a Van Gogh landscape style. TAV, CAMEL, SimDA as well as RAVE fails to realize the effects. CCEdit yields unclear styles between realistic and Van Gogh effect. In contrast, DAPE successfully produces a consistent and appealing Van Gogh style with specific visual elements. In the third column, the goal is to replace a squirrel with a rabbit. TAV struggles with facial consistency, CAMEL generates coarse details, and SimDA fails to maintain body shape. RAVE's motion is natural but lacks local detail, and CCEdit mistakenly creates a rabbit-squirrel hybrid. Conversely, DAPE successfully preserves consistent rabbit characteristics and generates high-quality details with clear semantics.
In short, qualitative results indicate that our proposed DAPE method outperforms the baselines in terms of temporal consistency, text alignment and detail fidelity, ultimately leading to noticeably improved visual smoothness and semantic relevance in the edited video.


\noindent \textbf{User Study.}
While CLIP-F and CLIP-T provide useful evaluations, they cannot fully account for human perceptual judgments~\cite{zhang2018perceptual}. Therefore, we conduct a user study to further validate our method. A total of 1,536 responses were collected from 30 participants, each completing a questionnaire with 25 sets of comparisons. Participants are asked to rank top-three videos (best, second-best, third) based on the following three criteria: 1) Which video aligns best with the textual description? 2) Which video is the smoothest, with the least local distortions and flickering? 3) Which video appears most visually refined overall? As in Figure~\ref{fig:userstudy}, our method outperforms the baselines, illustrating better human intuition following, temporal continuity and visual fineness. An example of our questionnaire is provided in the supplementary materials.

\subsection{Ablation Study}
\label{Ablation Study}

In this section, we conduct ablation experiments on two key issues of DAPE: one is the embedding location of the adapter in the second stage, and the other is the ablation of its internal design.
\begin{figure}[h]
    \centering
    \includegraphics[width=.8\columnwidth]{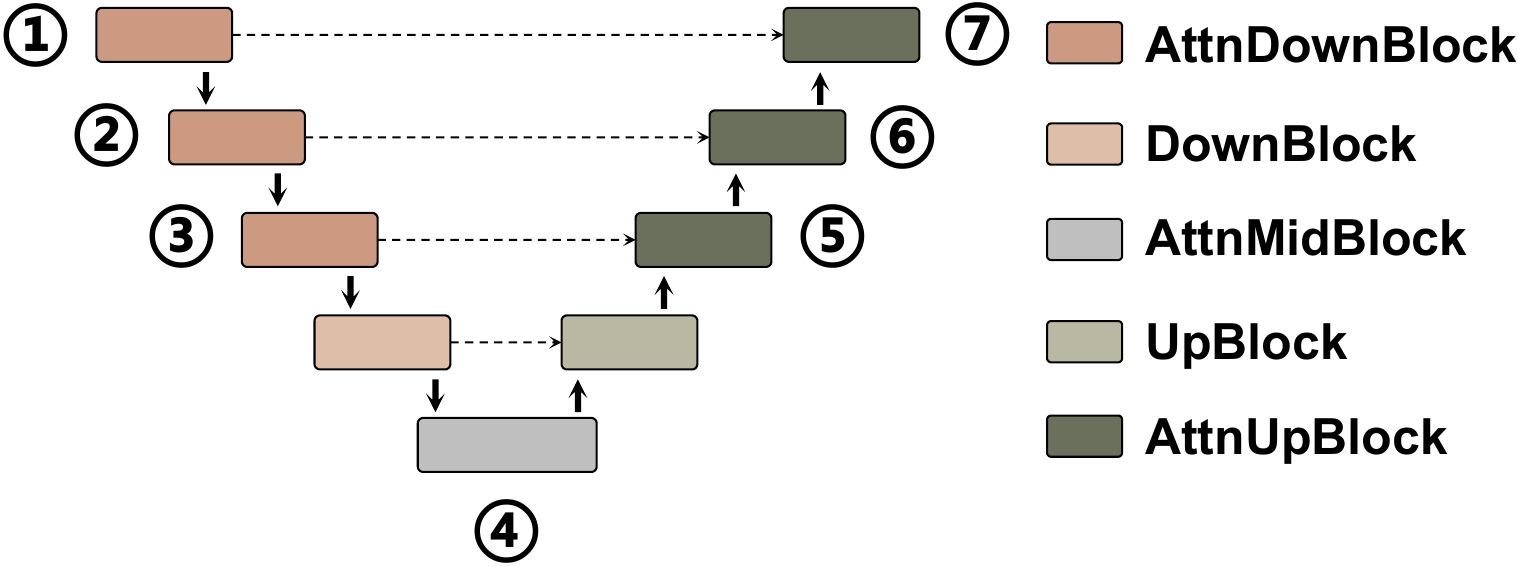}
    \caption{\textit{Ablation of adapters}. For better clarity, we index UNet blocks from \ding{172} to \ding{178}. The results are shown in Table \ref{tab:adapter_ablation}.}
    \label{fig:UnetArchi}
\end{figure}

\begin{figure}
    \centering
    \includegraphics[width=0.9\columnwidth]{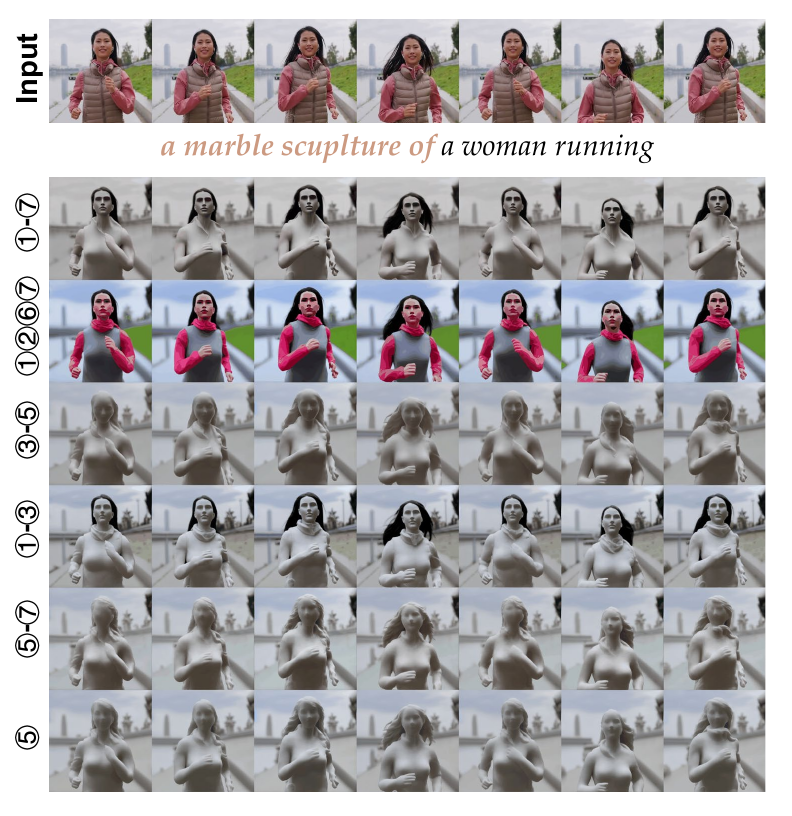}
    \caption{\textit{Illustration of adapter ablation}. The editing prompt requires changing the visual style to a marble sculpture. \ding{172}--\ding{178}, \ding{172}\ding{173}\ding{177}\ding{178}, and \ding{172}--\ding{174} fail to effectively follow the editing instruction. \ding{174}--\ding{176} negatively impact the facial lighting details, while \ding{176}--\ding{178} struggle to maintain temporal consistency. \ding{176} achieves the optimal editing results.}
    \label{fig:ablationstudy}
\end{figure}

\noindent \textbf{Adapter Position.}
We evaluate the impact of adapter placement by testing different positions within the U-Net’s attention blocks, labeled from \ding{172} to \ding{178} (Figure~\ref{fig:UnetArchi}). As shown in Table~\ref{tab:adapter_ablation}, inserting adapters at all positions (\ding{172}-\ding{178}) leads to degraded performance, especially in temporal consistency and alignment, likely due to overfitting and interference with low-level features. Inserting in shallow layers (\ding{172}\ding{173}\ding{177}\ding{178}) improves semantic consistency (highest CLIP-F) but results in poor structural coherence (highest Int.Err. and lowest Int.PSNR), suggesting a trade-off between semantic modeling and smooth generation. Deep layers placements (\ding{174}-\ding{176}) achieve better balance but still involve unnecessary complexity and attention redundancy. Placing the adapter only at the first block of the decoder yields the best overall results(\ding{176}). It achieves better smooth generation, decent semantic coherence as well as text alignment, indicating effective semantic reconstruction without disrupting earlier feature encoding. We finally adopt this setup to achieve enhancement of visual feature understanding. Figure~\ref{fig:ablationstudy} shows an example among six settings.

\begin{table}[htbp]
\centering
\scalebox{.9}{\setlength{\tabcolsep}{1mm}{
\begin{tabular}{l|cccc|c}
\hline
\multirow{3}{*}{\textbf{Method}} & \multicolumn{4}{c|}{\textbf{Temporal Consistency}} & \textbf{Alignment} \\
\cline{2-6}
 & CLIP-F $\uparrow$ & Int. Err. $\downarrow$ & Int. PSNR $\uparrow$ & War. Err. $\downarrow$ & CLIP-T $\uparrow$ \\
 & \(\times 10^{-2}\) & \(\times 10^{-2}\) & \(\times 1\) & \(\times 10^{-2}\) & \(\times 10^{-2}\) \\
\hline

\ding{172}-\ding{178}  & 94.59 & 9.10 & 21.57 & 2.71 & 28.66 \\
\ding{172}\ding{173}\ding{177}\ding{178}  & \textbf{94.88} & 9.41 & 21.41 & 3.00 & 29.00 \\
\ding{174}-\ding{176}         & 94.78 & 8.69 & \underline{21.96} & \underline{2.55} & \underline{29.52} \\
\ding{172}-\ding{174}         & 94.78 & 9.18 & 21.69 & 2.94 & 29.02 \\
\ding{176}-\ding{178}           & 94.68 & \underline{8.68} & 21.93 & \underline{2.55} & 29.34 \\
\ding{176}   & \underline{94.81} & \textbf{8.62}& \textbf{21.98} & \textbf{2.50} & \textbf{29.57} \\
\hline
\end{tabular}%
}}
\caption{\textit{Ablation Results of Adapter Position.} \ding{176} is selected as the final setting due to its balance of two judging dimensions.}
\label{tab:adapter_ablation}
\vspace{-15pt}
\end{table}

\noindent \textbf{Module Impact.}
We conduct comprehensive ablation experiments to evaluate the contributions of individual components within our approach. Specifically, we add adjustable normalization (A. N.) and visual adapter (V. A.) respectively, and try to train both modules simultaneously or adopt a dual-stage method. From Table~\ref{tab:module_ablation}, despite the improvement of the quality of fine visual details and inter-frame smoothness (lowest Int. Err and 
highest Int. PSNR), the visual adapter tends to reduce the temporal consistency and text alignment of generate videos. In contrast, the adjustable normalization contribute more significantly to maintaining consistent semantic representations across frames and improving text-image alignment. By combining these two modules, we find that training the normalization layer and the adapter simultaneously creates an negative interaction and compromise their respective strengths, while using dual-stage training strategy helps to relieve the mutual negative influence and even achieves better performance on War. Err. In short, our proposed framework employs dual-stage parameter-efficient fine-tuning methods, integrating adjustable norm tuning and visual adapter, to achieve a balanced trade-off among temporal consistency, human intention alignment, and detailed quality generation, leading to satisfying overall performance.

\begin{table}
\centering
\scalebox{.9}{\setlength{\tabcolsep}{1mm}{
\begin{tabular}{l|cccc|c}
\hline
\multirow{3}{*}{\textbf{Method}} & \multicolumn{4}{c|}{\textbf{Temporal Consistency}} & \textbf{Alignment} \\
\cline{2-6}
 & CLIP-F $\uparrow$ & Int. Err. $\downarrow$ & Int. PSNR $\uparrow$ & War. Err. $\downarrow$ & CLIP-T $\uparrow$ \\
 & \(\times 10^{-2}\) & \(\times 10^{-2}\) & \(\times 1\) & \(\times 10^{-2}\) & \(\times 10^{-2}\) \\
\hline
w/o All  & 94.85 & 8.71 & 21.94 & \multicolumn{1}{c|}{2.53} & 29.76 \\
w V. A.    & 94.71 & \textbf{8.25} & \textbf{22.58} & \underline{2.47} & 29.34 \\
w A. N.    & \textbf{95.05} & 8.69 & 22.01 & 2.62 & \textbf{29.82} \\
One-stage & 94.76 & 8.37 & \underline{22.51} & 2.50 & 29.42 \\
Ours        & \underline{94.98} & \underline{8.34} & 22.30 & \textbf{2.42} & \underline{29.78} \\
\hline
\end{tabular}%
}}
\caption{\textit{Ablation on inner design of DAPE.} The proposed two-stage setting can outperform baseline on all metrics.}
\label{tab:module_ablation}
\vspace{-20pt}
\end{table}


\section{Conclusion}

In this paper, we introduce DAPE, a dual-stage parameter-efficient fine-tuning framework with adjustable norm-tuning and a carefully positioned visual adapter, to significantly enhance the temporal consistency and visual quality and generate more consistent videos. Accompanying this framework, we propose DAPE Dataset, a comprehensive benchmark designed to systematically evaluate performance across diverse editing scenarios. Extensive experimental validation confirmed that our approach achieves state-of-the-art results, effectively balancing visual quality, temporal coherence, and prompt adherence, paving the way for future research in generative model optimization and broader applications.

\bibliographystyle{ACM-Reference-Format}
\bibliography{main}

\clearpage
\appendix              
\section{Dataset}
\begin{figure}
    \centering
    \includegraphics[width=0.47\textwidth]{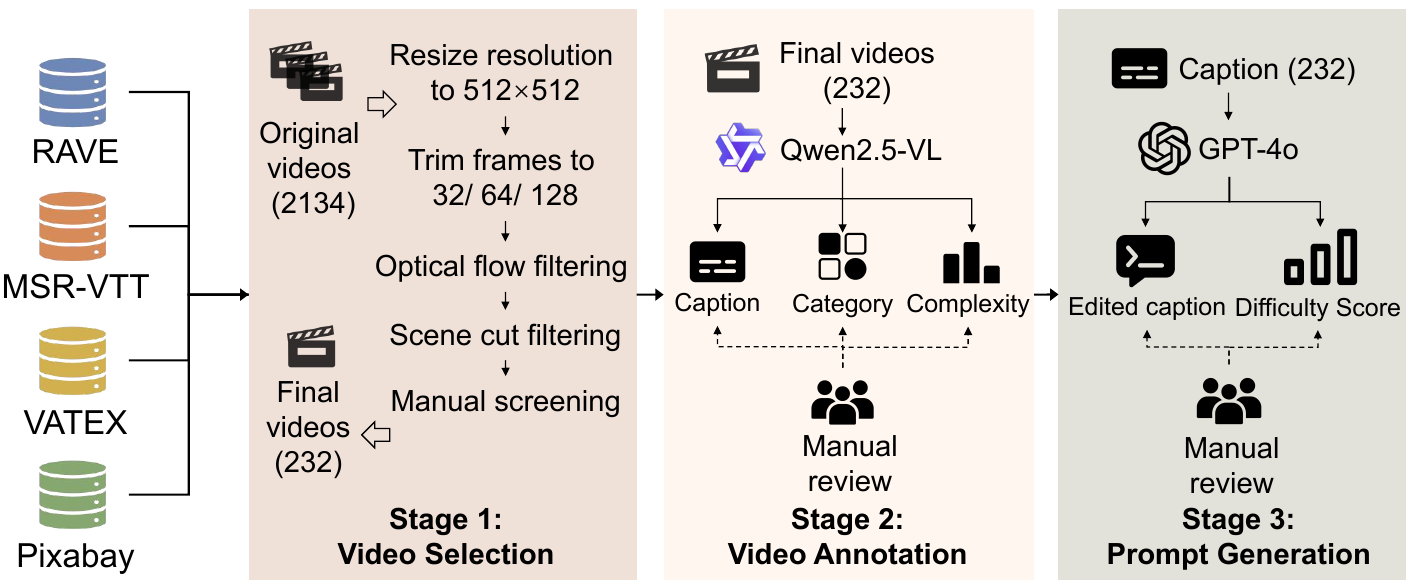}
    \caption{Overview of the three-step pipeline for dataset construction}
    \label{fig:datasetPipeline}
\end{figure}

\begin{figure*}
    \centering
    \includegraphics[width=0.9\textwidth]{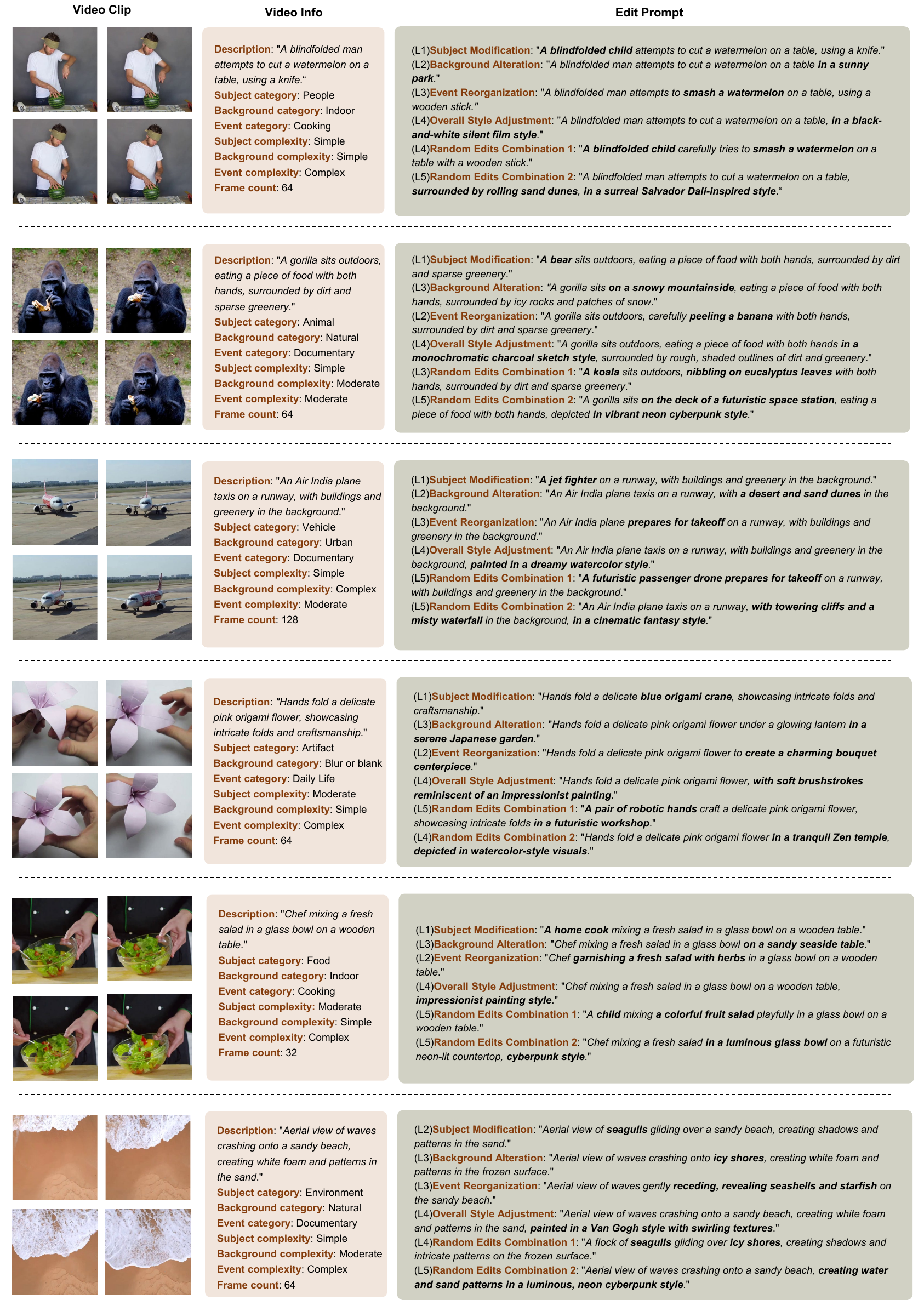}
    \caption{Illustrative examples of our DAPE Dataset. The labels (L1–L5) indicate the difficulty levels of the editing tasks.}
    \label{fig:datasetIllustration}
\end{figure*}

The construction of our dataset is organized into three sequential steps: video selection, video annotation, and prompt generation, as illustrated in Figure ~\ref{fig:datasetPipeline}. This pipeline is specifically designed for building video editing datasets, integrating both automated tools and human validation. Our DAPE Dataset consists of 232 videos, each accompanied by detailed annotations and multiple prompts for video editing tasks, as illustrated in Figure ~\ref{fig:datasetIllustration}.

\textbf{Video Source. }We initially collected 2,134 videos from four sources, including RAVE~\cite{kara2024rave}, MSR-VTT~\cite{xu2016msr}, VATEX~\cite{wang2019vatex}, and Pixabay~\cite{pixabay}. For large-scale datasets such as VATEX, random sampling was applied to reduce redundancy. Videos with resolutions below 512×512 were discarded, and the remaining ones were resized to 512×512 and trimmed to 32, 64, or 128 frames. We then apply optical flow filtering to exclude samples with excessive motion and scene cut filtering to remove videos with abrupt transitions, similar to the strategy used in LVD-F~\cite{blattmann2023stable} and EgoVid~\cite{wang2024egovid}. Finally, a manual screening process was conducted to ensure quality, resulting in a curated set of 232 high-quality videos.

\begin{figure*}
    \centering
    \includegraphics[width=0.85\textwidth]{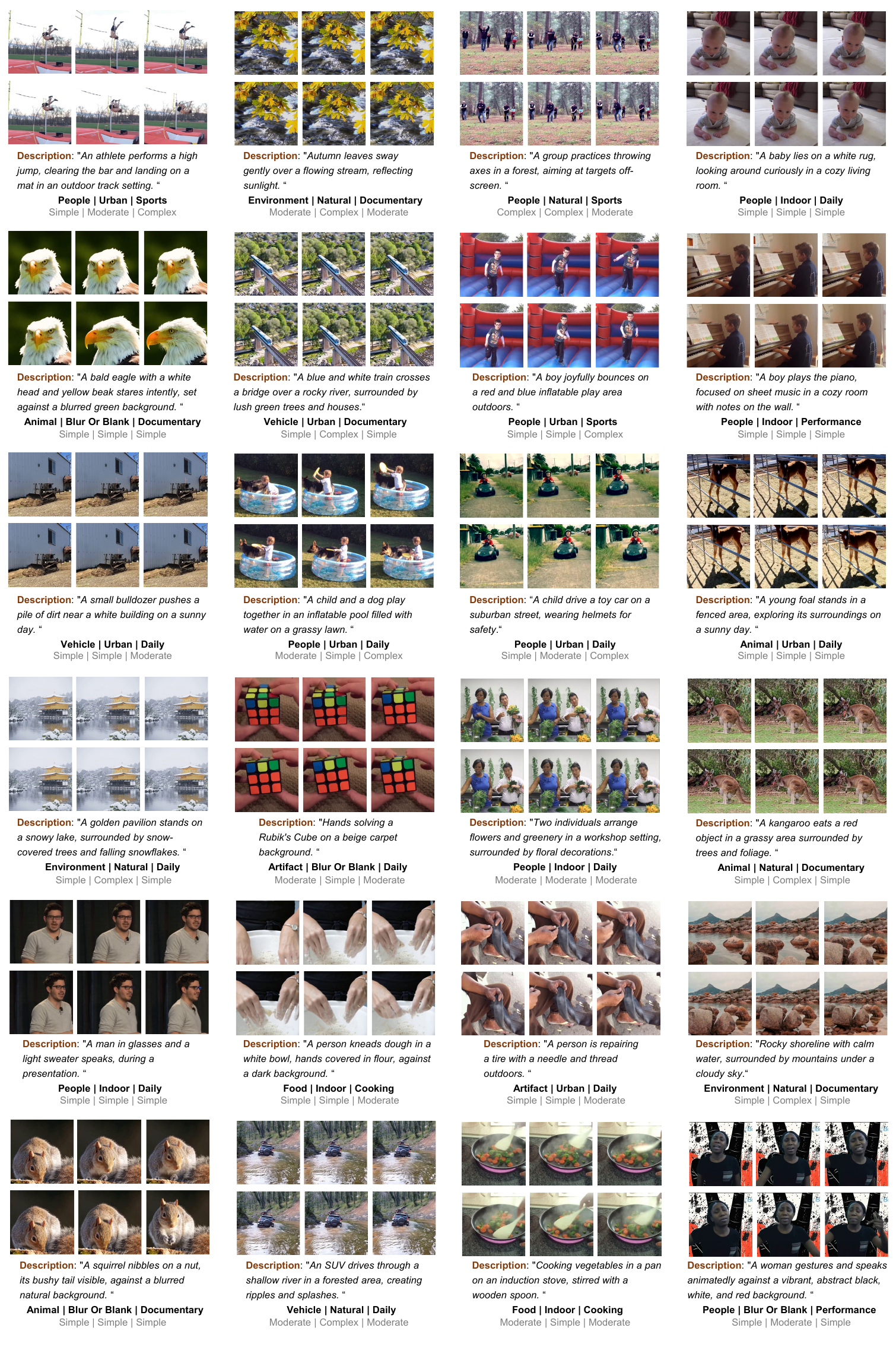}
    \caption{More sample video frames from our DAPE Dataset}
    \label{fig:datasetExample}
\end{figure*}

\textbf{Video Annotation. }For each video, we provide a sequence of eight evenly spaced frames to the Qwen2.5 vision-language model, which is capable of capturing temporal context and understanding video-level semantics from the frame sequence. The model was instructed to generate three types of annotations for each video: a caption, semantic category labels (subject, background, and event), and complexity scores for each component. All automatically generated annotations are then manually reviewed and corrected to ensure semantic accuracy and consistency. Detailed examples of the videos and their corresponding annotations are illustrated in Figure~\ref{fig:datasetExample}.

\textbf{Prompt Generation. }The verified captions are passed to the GPT-4o model, which is prompted to generate editing tasks from five perspectives: subject modification, background alteration, event reorganization, overall style adjustment, and random combinations thereof. For each task, the model returns a revised edited caption and a corresponding difficulty score. All outputs are further manually reviewed to ensure clarity, feasibility, and correctness.

\textbf{Video Selection Criteria. }After automated filtering, all candidate videos underwent a round of manual quality screening. A video was considered acceptable if it satisfied the following conditions across four key dimensions.
\begin{itemize}
    \item \textbf{Motion}: Both camera and subject movement should be smooth and stable, without abrupt shaking or erratic motion.
    \item \textbf{Editing:} The video should maintain temporal continuity, with no scene cuts, montage transitions, or long static frames.
    \item \textbf{Content:} The visual subject must be complete and unobstructed, with no prominent overlaid text or distracting visual elements.
    \item \textbf{Visual Quality:} The overall presentation should be aesthetically coherent, with appropriate lighting, contrast, and composition.
\end{itemize}

\textbf{Video Categories. }Each video in our dataset is categorized based on three core components—subject, background, and event. The specific category sets for each component are adapted from the classification scheme used in MSR-VTT~\cite{xu2016msr}, with modifications to better suit our video editing context.
\begin{itemize}
    \item \textbf{Subject: }Indicates the primary entity or focus present in the video, including \textit{people}, \textit{animal}, \textit{vehicle}, \textit{artifact}, \textit{food} and \textit{environment}.
    \item \textbf{Background:} Describes the dominant scene or setting in which the video takes place, including \textit{indoor}, \textit{urban}, \textit{natural} and \textit{blur or blank}.
    \item \textbf{Event:} Refers to the main activity or situation depicted in the video, including \textit{sports}, \textit{daily}, \textit{performance}, \textit{documentary} and \textit{cooking}.
\end{itemize}

\textbf{Types of Edits. }Each video in our dataset is associated with five types of editing tasks, each targeting different aspects of the video content.
\begin{itemize}
    \item \textbf{Subject Modification:} Alters the appearance or identity of the primary subject in the video such as \textit{changing clothing}, \textit{replacing a person with an animal}.
    \item \textbf{Background Alteration:} Modifies the visual setting or environment in which the video takes place such as \textit{changing a kitchen scene to a grassland}.
    \item \textbf{Event Reorganization:} Modifies the main action or activity depicted in the video such as \textit{changing a person walking a dog to playing basketball}.
    \item \textbf{Overall Style Adjustment:} Changes the visual tone or artistic style of the video such as \textit{applying cartoon effects},  \textit{converting to black-and-white}.
    \item \textbf{Random Edits Combination:} Randomly applies a combination of two editing types selected from the four categories above.
\end{itemize}

\begin{figure*}
    \centering
    \includegraphics[width=0.9\textwidth]{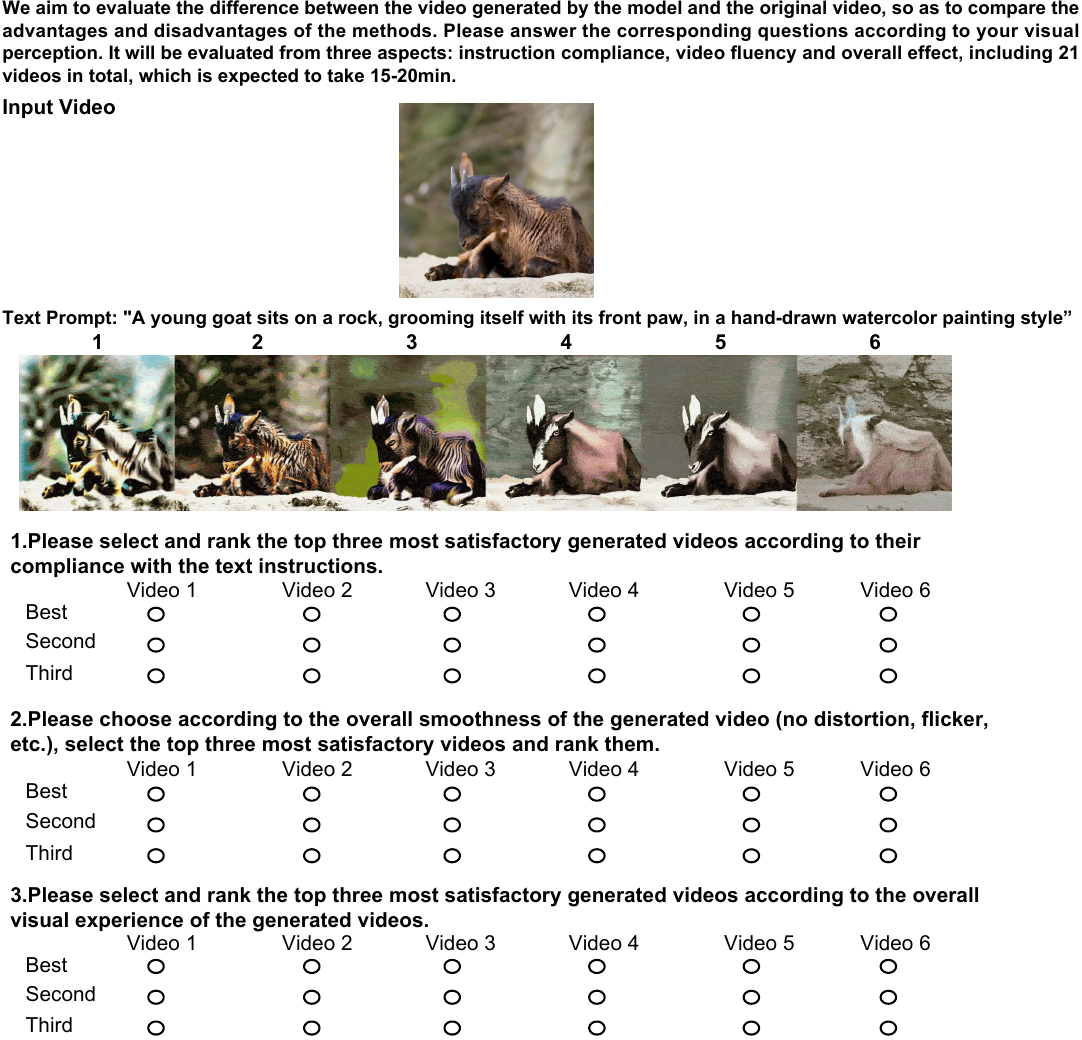}
    \caption{Questionnaire example of user study.}
    \label{fig:userStudy}
\end{figure*}

\section{User Study}

We conducted a user study by recruiting anonymous participants. The study fo-cused on 21 randomly selected video-text pairs from our dataset. The comparison in the user study was made among CCEdit, RAVE, SimDA, CAMEL, TAV, as well as our DAPE approach. Notably, we focused on questions related to textual alignment, temporal consistency, and general editing capabilities. Figure \ref{fig:userStudy} shows the questionnaire example.

\end{document}